\pdfoutput=1
\documentclass[11pt]{article}

\usepackage{acl}

\usepackage{times}
\usepackage{latexsym}

\usepackage[T1]{fontenc}
\usepackage[utf8]{inputenc}

\usepackage{microtype}

\usepackage{inconsolata}

\usepackage{graphicx}

\usepackage{tikz-dependency}
\tikzset{/depgraph/.cd,/depgraph/.search also = {/tikz},
baseline=-0.6ex, inner sep=-0.1cm, edge horizontal padding=3pt, edge unit distance=1.8ex}
\usepackage{gb4e}
\noautomath

\title{Dependency Annotation of Ottoman Turkish with Multilingual BERT}

\author{Şaziye Betül Özateş$^{\ast}$, 
   Tarık Emre Tıraş$^{\ast}$,
       Efe Eren Genç$^{\ast}$, \\
 \bf  Esma F. Bilgin Taşdemir$^{\ddagger}$
  \\
  $^{\ast}$Boğaziçi University,  $^{\ddagger}$Medeniyet University, Turkey \\
   \texttt{saziye.ozates@bogazici.edu.tr}, \texttt{\{tarik.tiras,efe.genc\}@std.bogazici.edu.tr}, \\\texttt{esmabilgin.tasdemir@medeniyet.edu.tr}
 \\}

\begin{document}
\maketitle
\begin{abstract}
 This study introduces a pretrained large language model-based annotation methodology for the first dependency treebank in Ottoman Turkish.
Our experimental results show that, iteratively, i) pseudo-annotating data using a multilingual BERT-based parsing model, ii) manually correcting the pseudo-annotations, and iii) fine-tuning the parsing model
 with the corrected annotations, we speed up and simplify the challenging dependency annotation process. The resulting treebank, that will be a part of the Universal Dependencies (UD) project, will facilitate automated analysis of Ottoman Turkish documents, unlocking the linguistic richness embedded in this historical heritage.
\end{abstract}

\section{Introduction}
The Ottoman Turkish is one of the historical versions of Turkish. It was in use for nearly six centuries before it evolved into modern Turkish. During that period, it went through many changes in terms of lexicon and, to some extent, syntax.

There is an increasing trend in digitization campaigns that aim to preserve historical documents and facilitate user access to these resources. Numerous institutions publicly share images of their historical documents written in Ottoman Turkish. As these historical materials become more accessible in digital form, a crucial need for automated processing and retrieval of their content arises.
However, unlike modern languages with well-established linguistic resources and corpora, Ottoman Turkish has quite a limited availability of annotated data, dictionaries, and linguistic references. That scarcity of resources hinders the development and training of robust NLP models for the Ottoman Turkish.

A possible solution for the problem of resource scarcity can be leveraging Turkish resources based on the fact that they are actually two versions of the same language. However, the linguistic evolution from Ottoman Turkish to modern Turkish involves substantial changes over time. This transition impacts semantic nuances, vocabulary, and grammatical structures, making it challenging to apply contemporary Turkish NLP techniques directly to historical texts.

It is a common approach to prepare the necessary resources manually in the case of low-resource languages. However, manual tagging and annotation are both time-consuming and expensive. In the case of dependency annotation, there is also a need for experts specialized in the task at hand.

To simplify and speed up the manual annotation process for our objective of constructing an Ottoman Turkish dependency treebank, we propose to use pretrained Large Language Models (LLMs) as pseudo-annotators. LLMs have proven themselves as powerful tools widely used in various fields due to their adaptability. They excel in understanding natural language, making them valuable across diverse areas, from downstream tasks like machine translation \cite{yang2020towards} and sentiment analysis \cite{hoang2019aspect} to data annotation \cite{tejani2022performance}. We experiment with a multilingual BERT-based parsing model for dependency annotation in our treebank. Our proposed methodology suggests generating pseudo-annotations using the parsing model, then correcting them manually and fine-tuning the model with the corrections, and then using the updated model to pseudo-annotate new data in a repeated cycle.

In this study, we present our LLM-enhanced syntactic annotation methodology for the first Ottoman Turkish dependency treebank, which will be a part of the Universal Dependencies (UD) project \cite{nivre2016universal}. We investigate the impact of enhancing the accuracy of pseudo-annotations by fine-tuning the parsing model with labeled Ottoman Turkish data. We show that utilizing a multilingual BERT-based parsing model in dependency annotation leads to better pseudo-annotations, which eases the manual annotation of the Ottoman Turkish treebank job for human annotators. With the creation of this treebank, we aim at facilitating the automatic analysis and understanding of Ottoman Turkish documents, thereby facilitating a more comprehensive exploration of this linguistically rich heritage.

\section{Background and Related work}

We present some background on the Ottoman Turkish
and discuss its differences from modern Turkish in this section. The section includes a summary of the related works from the literature as well.

%We discuss the challenges encountered in annotating Ottoman Turkish texts.

\subsection{The Ottoman Turkish}
Ottoman Turkish is the historical form of the Turkish language used in the Ottoman Empire from the late 13th to the early 20th century.
%It initially used the Arabic script, later transitioning to the Perso-Arabic script. 
It has a writing system based on an extended version of the Arabic script where five additional characters are used to represent some Turkish sounds. Ottoman Turkish has a significant number of loanwords borrowed from Arabic and Persian.
Furthermore, it contains some complex grammatical forms borrowed from these two languages, like long noun phrases with multiple words and particular word ordering. Finally, the number of words in a sentence can get quite large, especially in the texts written in the early periods of the Ottoman Empire.

The Arabic alphabet-based writing system was replaced with a Latin alphabet-based one in 1928. Also, the language reforms in the early 20th century led to significant modifications in the vocabulary, where many of the loanwords of Arabic and Persian origin were replaced with words borrowed from European languages or newly coined words. These language reforms have led to a substantial divergence between modern and Ottoman Turkish.

A comparison of modern Turkish and Ottoman Turkish in terms of the percentage of loanwords can be helpful in understanding the differences between the two versions. The well-known Ottoman dictionary "A Turkish and English Lexicon" by James W. Redhouse \cite{Redhouse:1884} contains 79,491 words, of which 57\% are Arabic-origin and 12\% are Persian-origin. In contrast, a study by \cite{Moore:2015} reports that approximately 25\% of the selected 3,270 most frequently used Turkish words are Arabic-origin, and the percentage of Persian loanwords is only 6\% in the selected vocabulary. On the other hand, of the 104.481 words in the standard Turkish Dictionary by TDK (Turkish Language Institution), 6,463 (\textasciitilde6\%) are Arabic-origin and 1,374 (\textasciitilde1\%) are Persian-origin \cite{TDK:2005}.

\subsection{Related Work}

The Universal Dependencies (UD) Project\footnote{\url{https://universaldependencies.org/introduction.html}} is a collaborative effort that aims to develop cross-linguistically consistent treebank annotation for many languages. The project provides a standardized framework for annotating the grammatical structure of sentences, encompassing parts of speech, syntactic dependencies, and other linguistic features.
Most of the treebanks in UD are dedicated to annotating modern languages. Efforts to include historical languages in the project have been limited. Ancient Greek is one of the lucky historical languages having three UD treebanks \cite{bamman2011ancient}. Ancient Hebrew \cite{swanson2022universal}, Classical Chinese \cite{lee2012dependency}, and Coptic \cite{zeldes2018coptic} treebanks are some other notable UD treebanks that contain annotated sentences from historical languages.
Using trained models as pseudo-annotators is a common practice in data annotation \cite{haunss2020integrating,benato2021iterative,he2022generate}. In the context of dependency annotation, there are some studies that utilize automatic parsers to parse raw data as an initial step in order to provide human annotators with a starting point in the manual annotation process (e.g., the Belarusian HSE UD Treebank\footnote{\url{https://github.com/UniversalDependencies/UD_Belarusian-HSE}}). However, to the best of our knowledge, these studies use machine assistance as a one-time process only; none of them adopted an iterative approach as we do.

\section{OTA-BOUN: A UD Treebank for Ottoman Turkish }

As an initial endeavor towards enriching Ottoman Turkish Natural Language Processing (NLP) resources  and enhancing the effectiveness of existing NLP tools on Ottoman Turkish texts, we have initiated the creation of the Ottoman Turkish UD Treebank. In Section \ref{sec:data}, we provide information about the source and language of the treebank. Section \ref{sec:annot-scheme} details our annotation process, and Section \ref{sec:annot-challenge} states the challenges encountered in annotating Ottoman Turkish sentences, accompanied by illustrative examples, and clarifies the approaches undertaken to overcome these challenges. 

\subsection{Data}
\label{sec:data}
\begin{figure*}[h]

\centering
\includegraphics[width=\textwidth]{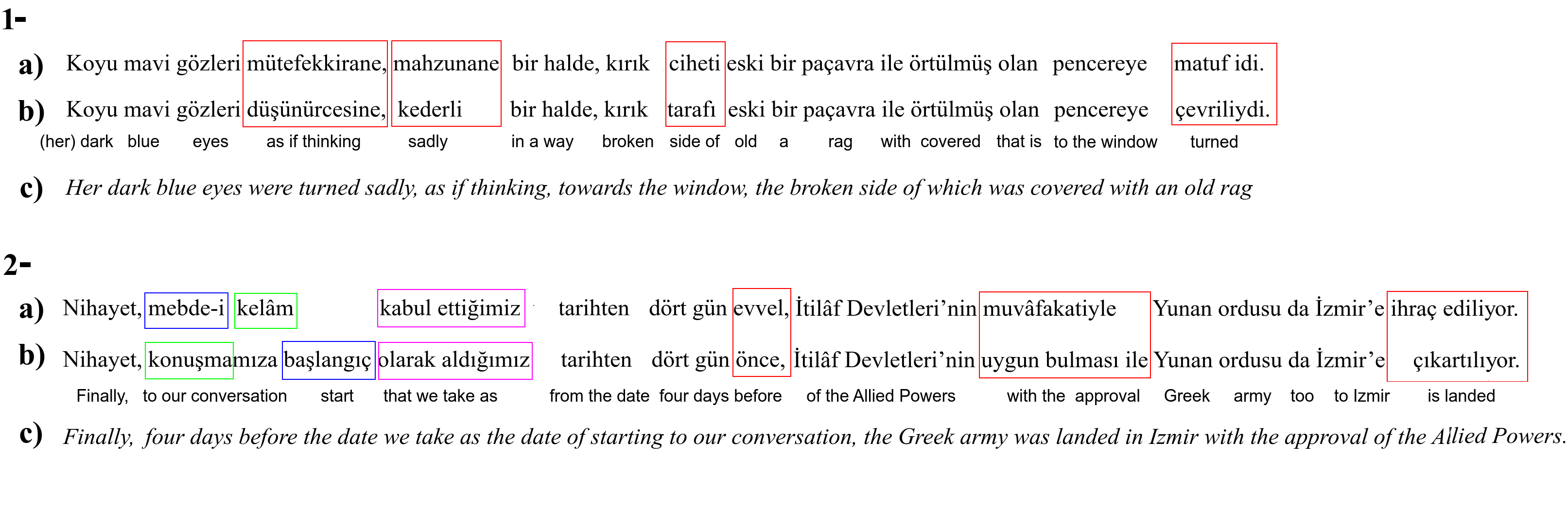}
\caption{Two Ottoman Turkish sentences (in Latin script), their modernized versions and English translations.}
\label{fig_sentences}
\end{figure*}

The creation of an Ottoman Turkish dependency treebank is an ongoing project. The first version of the treebank\footnote{The treebank will be released with the next scheduled release of UD (v2.14).} is planned to contain 500 syntactically annotated Ottoman Turkish sentences. The annotated sentences exist in two writing styles in the treebank; 1) written with the Latin-based Turkish alphabet, and 2) written with the Perso-Arabic alphabet. (See Figure \ref{fig_ota_boun} for CoNLL-U representation\footnote{\url{https://universaldependencies.org/format.html}} of treebank items.)

The sentences are sampled from seven texts by four different writers. All of the texts are from literature published between 1900 and 1928. There are two articles, excerpts from two history texts, two stories, and one excerpt from a novel. Our ultimate goal is to include annotations for a total of 2,000 sentences, sourced from various origins, in the final version.

Figure \ref{fig_sentences} shows two example sentences from the dataset in Ottoman and modern Turkish.In the figure, some of the words that have been replaced by modern versions can be seen in red boxes. A group of defunct words can be replaced by a single word (ex. \textit{ ihraç ediliyor --> çıkarılıyor (is being extracted)}) and vice versa (ex. \textit{muvafakatiyle --> uygun bulması ile (with the approval of)}). A noun phrase form that is mostly obsolete is replaced with its modern counterpart where the word order changes (ex. \textit{mebde-i kelam --> konuşma başlangıç(ı) (the beginning of the speech)}). Here, the words are also replaced with synonyms in the modern version.

\begin{figure*}[!htb]
\centering
\includegraphics[scale=0.9]{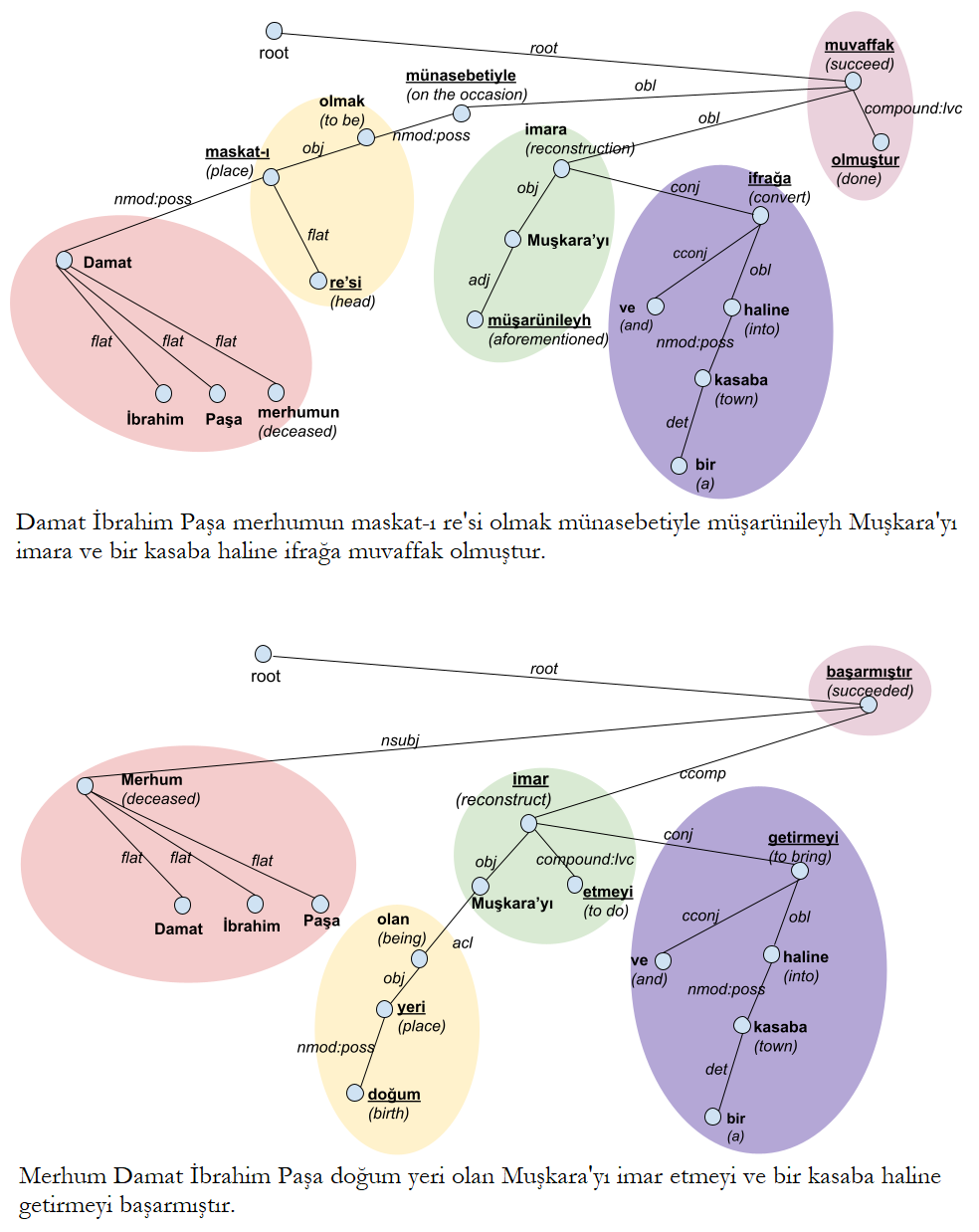} 
\caption{ Dependency tree representations of an Ottoman Turkish sentence (above) and its rephrased version using the modern Turkish (below). The highlighted portions enclosed in colored circles indicate corresponding segments in the sentences. English translations of words are provided in italics within parentheses. Words of a sentence that do not exist in the other sentence are underlined in the figure. {\textbf{English translation of the sentence:}} "The late Damat İbrahim Paşa succeeded in developing Muşkara, his birthplace, and turning it into a town."}
\label{fig_deptree}
\end{figure*}

In addition to changes in vocabulary, the syntactical features of an Ottoman Turkish sentence can be different from those of the modern version. Figure \ref{fig_deptree} shows the dependency trees of two versions of a sentence: the original one and its translation to modern Turkish. The difference in grammatical structure of both sentences is visible through their dependency trees. To give a concrete example, the way the subject of the sentence ({\it Damat İbrahim Paşa}), which is a person, is connected to the verb is different in the original sentence (via an {\textit{oblique}} relation through one of its parent nodes) and in its translation (directly via {\textit{nominal subject}} relation).

Another example of their differences is in the way they indicate that {\it Muşkara}, a place name, is the subject's birthplace. This is done indirectly in the original sentence by connecting them at the {\textit{root}} word. The connection is more straightforward in its translation to modern Turkish, which is done directly using a clausal modifier.Due to such differences in the grammar and vocabulary of Ottoman and modern Turkish, the annotation process is not straightforward.

\subsection{The Annotation Scheme}
\label{sec:annot-scheme}

\begin{figure*}[h]

\centering
\fbox{\includegraphics[width=0.95\textwidth]{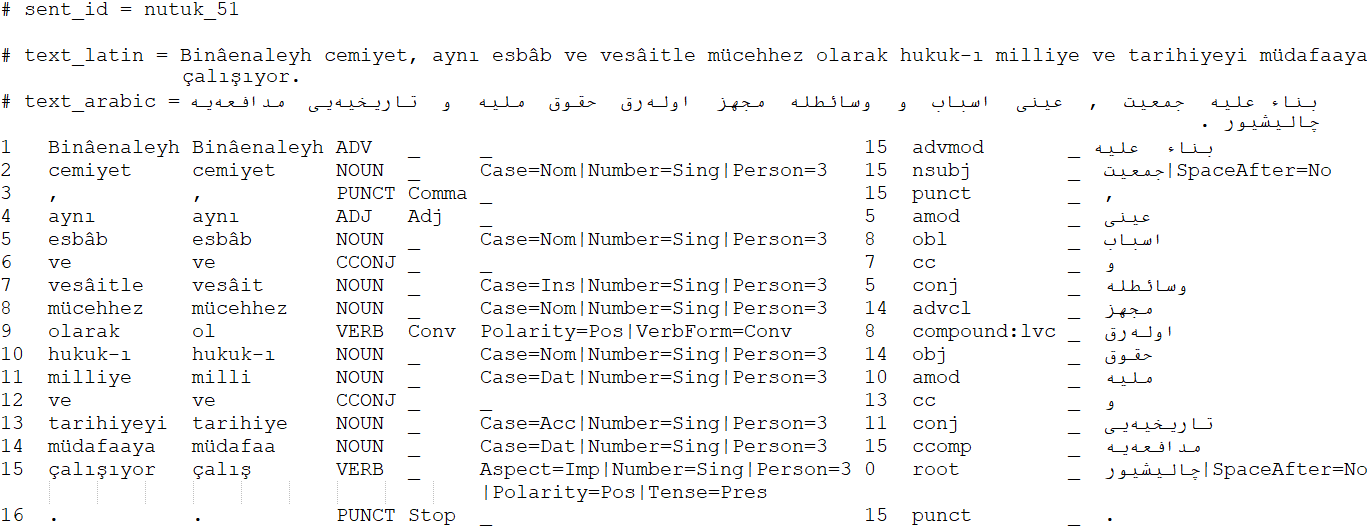}}
\caption{CoNLL-U Representation of an example sentence from our Ottoman BOUN UD Treebank.}
\label{fig_ota_boun}
\end{figure*}

For the annotation of the Ottoman Turkish treebank, we used a team of two annotators who are linguists and
have comprehensive knowledge of Turkish grammar and general linguistics, as well as grammatical theories. Assisting the team of annotators in the process of manual annotation are two senior computer scientists with extensive expertise both in NLP and in Ottoman Turkish.

In the annotation process, we employ double annotation for a randomly chosen set of 50 sentences. We measure Cohen's kappa between the annotated dependency labels as 0.85 for determining the inter-annotator agreement. The unlabelled and labeled attachment scores between the annotations are found to be 82.20\% and 76.91\%, respectively. The rest of the annotation is held separately. After each annotator finishes their respective portions, the annotated sentences are examined, and any disagreement is resolved within the linguist team.

In most cases, we adhered to the conventions of Universal Dependencies but referred to the Suggested UD Guidelines for Turkish\footnote{\url{https://github.com/boun-tabi/UD_docs/blob/main/_tr/dep/Turkish_deprel_guidelines.pdf}} when necessary. At the current step, we do the manual annotation on the syntactic level. Following the UD scheme, the annotated data is stored in CoNLL-U format. Figure \ref{fig_ota_boun} depicts this format on an annotated sentence from our treebank. As mentioned, we preserve the original Arabic script version of the data, providing the text value both in Latin and Arabic letters. The tokens of the sentence in Latin script are found in the second column, while their Arabic counterparts are provided in the last column.

\subsection{Challenges in the Syntactic Annotation}
\label{sec:annot-challenge}

\paragraph{Deformation of Turkish Morphosyntax}

In the process of annotating Ottoman Turkish, a significant challenge encountered was that certain structures influenced by Arabic and Persian do not require Turkish inflectional suffixes, which distinguishes them from modern Turkish syntax and morphology. For example, \textit{münasebet tabian} in Example \ref{ex:münasebet} means \textit{subject to relationship}. The word \textit{tabian} is a verbal adverb derived from the Arabic word \textit{tabi (subject to)} and the Arabic adverbial suffix \textit{-An}. In modern Turkish, the phrase can be expressed as follows: \textit{münasebete tabi olarak} \textit{(by being subject to a relationship)}. As seen, the noun \textit{münasebet} is inflected with a dative case {\textit{-A}}. In Ottoman Turkish, however, a dative case is not needed in some instances, making a difference from modern Turkish. The excessive exposure to Arabic led some suffixes to disappear without a narrowing in meaning, consequently causing difficulties in the annotation process.

\begin{exe}
\small
    \ex \label{ex:münasebet}
    \gll \it güneş-in \it sema \it üzerin-de-ki \it seyr-i \it ile \it {\color{blue}münasebet} \it {\color{blue}tabian}\\
    sun-\textsc{gen.3sg} sky above-\textsc{loc}-\textsc{adj} journey-\textsc{poss.3sg} \textsc{com} {\color{blue}relationship} {\color{blue}subject\_to}\\
    \glt `with respect to the trajectory of the sun in the sky'
\end{exe}
The application of Turkish suffixes to structures in Arabic and Persian poses difficulties for both the parser and our team. For instance, in the phrase {\textit{Harb-i Umumi'de (in the world war)}} in Example \ref{ex:harb}, although {\textit{harb}} is the head, the locative suffix is attached to the dependent, diverging from standard modern Turkish rules and leading to incorrect automatic annotations.
\begin{exe}
\small
    \ex \label{ex:harb}
    \gll \it Memleket-i \it {\color{blue}Harb-i} \it {\color{blue}Umumî-ye} \it sevk \it ed-en-ler\\
    country-\textsc{acc} {\color{blue}war-\textsc{ez}}  {\color{blue}general-\textsc{dat}} dispatch make-\textsc{rel}-\textsc{pl} \\
    \glt `Those who sent the country to the World War'
\end{exe}
\paragraph{Foreign Compounds}
Another challenge was identifying compound verbs formed with foreign words not commonly used in modern Turkish, such as {\textit{mehcur bırakmak (to leave abandoned)}}. The high frequency of uncommon phrasal verbs poses a challenge to the annotation process.

\paragraph{Abundance of Foreign Words}
It was additionally observed that the aforesaid cultural interactions influenced the lexicon of Ottoman Turkish. The influence led to confusion about determining the part-of-speech of certain words with two meanings without showing any structural difference. They make sense within the context in which they are used, making them challenging to annotate just by looking at their syntactic and morphological structures. In {\textit{sû-i muamelat (unpleasant behavior)}} given in Example \ref{ex:sui}, {\textit{sû}} is an adjective. However, in {\textit{sû-i tefehhüm (misunderstanding)}}, it acts as an adverb. At first glance, these phrases appear to have no difference, both syntactically and morphologically. This similarity poses a challenge, as it demands substantial lexical knowledge of Arabic.
\begin{exe}
\small
    \ex \label{ex:sui}
    \gll \it yap-ıl-an \it {\color{blue}su-i} \it {\color{blue}muâmel-ât-ta} \it millet-in \it medhaldar \it bul-un-ma-dığ-ın-ı\\
    do-\textsc{pass}-\textsc{rel} {\color{blue}bad-\textsc{ez}} {\color{blue}treatment-\textsc{pl}-\textsc{loc}} nation-\textsc{gen.3sg} responsible find-\textsc{pass}-\textsc{neg}-\textsc{nmlz}-\textsc{poss.3sg}-\textsc{acc}\\
    \glt `that the nation is not responsible for the mistreatment done'
\end{exe}
There are foreign words in Ottoman Turkish that preserve morphological features gained in their original language. The challenge arises from the fact that the derivational morphemes remain productive, making it difficult to determine their syntactic relations. The primary reason for this challenge is the necessity of understanding the nature of the language from which the words originate. Consider the phrase {\textit{i’tâ-yı ma’lûmat (giving information)}} in Example \ref{ex:i'tâ}. The verbal noun {\textit{i'tâ}}, derived from {\textit{aṭā (grant)}}, requires knowledge of Arabic grammar to identify it as a verbal noun. Such examples are prevalent in Ottoman Turkish, posing difficulties in annotation.
\begin{exe}
\small
    \ex \label{ex:i'tâ}
    \gll \it Hedef-i \it siyasi-ler-i \it hakkında \it {\color{blue}itâ-yı} \it {\color{blue}malum-ât} \it eyle-mek\\
    goal-\textsc{ez} political-\textsc{pl}-\textsc{poss.3} about {\color{blue}give.\textsc{nmlz}-\textsc{ez}} {\color{blue}known-\textsc{pl}} make-\textsc{nmlz}\\
    \glt `to give information about their political goals'
\end{exe}
\paragraph{Annotation of Regular Structures}
Another issue is annotating phrases borrowed from foreign languages. According to UD, names with a regular syntactic structure should be annotated with regular syntactic relations rather than being annotated as {\textit{flat}}. It is also noted that foreign language expressions should be annotated as {\textit{flat}}. Despite these guidelines, due to their significant effects, we aimed to preserve frequently used grammar structures obtained from foreign languages as much as possible. For instance, in {\textit{Harb-i Umumi (World War)}}, {\textit{harb}} is a noun meaning {\textit{war}} and {\textit{umumi}} is an adjective meaning {\textit{universal}}. This phrase could have been created in Turkish grammar as {\textit{Umumi Harp}}. Although the structure is foreign, we annotated {\textit{Harb-i Umumi}} and similar phrases as regular structures. 

%\paragraph{Omission of Genitive Case Marker}
%In Ottoman Turkish, we observed a more liberal omission of the genitive case in genitive-possessive constructions, complicating the parsing process in certain instances. 
%{\color{red}[EXAMPLE]}

\section{Iterative Annotation with Multilingual BERT}
\label{sec:annot-bert}

\begin{figure*}[h]

\centering
\fbox{\includegraphics[width=0.95\textwidth]{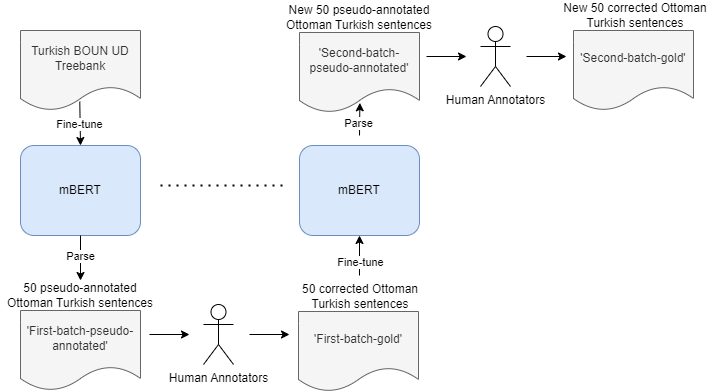}}
\caption{The experimental setup using the iterative annotation scheme.}
\label{fig_exper_setup}
\end{figure*}

Because dependency annotation is both cost- and time-intensive, and given the challenges outlined in the previous section, we recently contemplated whether we could ease our manual annotation phase by leveraging state-of-the-art Large Language Models (LLMs). To test this idea, we conducted an experiment where we utilized the multilingual BERT model\footnote{https://huggingface.co/bert-base-multilingual-cased} \cite{devlin2019bert}. The experiment is set up as follows:

\begin{itemize}
\item Initially, we fine-tuned a multilingual BERT-based parsing model \cite{grunewald2020applying} using the Turkish BOUN UD Treebank \cite{turk2022resources} containing only modern Turkish sentences.

\item The fine-tuned model pseudo-annotated a subset of 50 sentences from the Ottoman Turkish corpus. This pseudo-annotated 50 sentences will be referred to as `First-Batch-Pseudo-Annotated'.

\item Subsequently, our human annotators reviewed and corrected the `First-Batch-Pseudo-Annotated', the result of which is `First-Batch-Gold'.

\item We then further fine-tuned our parsing model with `First-Batch-Gold'.

\item Using this updated model, we pseudo-annotated another set of 50 sentences, denoted as the `Second-Batch-Pseudo-Annotated'.

\item Finally, our human annotators corrected `Second-Batch-Pseudo-Annotated', resulting in the `Second-Batch-Gold'.
\end{itemize}

Figure \ref{fig_exper_setup} depicts this iterative annotation scheme for two iterations.

%Initially, we fine-tune  a multilingual BERT-based dependency parser \cite{grunewald2020applying} on the Turkish BOUN Treebank \cite{turk2022resources} (henceforth referred to as BOUN). BOUN includes 9,761 sentences written in modern Turkish. Then, we automatically parse 50 Ottoman Turkish sentences from our corpus using the fine-tuned parser. Subsequently, two annotators manually reviewed these pseudo-annotated sentences and corrected the errors made by the parser. Then, we further fine-tune the parser using this corrected annotation of 50 sentences. Finally, we automatically parsed another batch of 50 Ottoman Turkish sentences by using this updated parsing model. 

Thus, in this initial phase of annotation, we have 100 manually annotated Ottoman Turkish sentences in two iterative batches of 50 sentences each. We should note that, because of the difficulty in finding parallel Ottoman Turkish and modern Turkish data, the sentences in these two batches originated from different sources and different genres. This has led to variations in the annotation complexity of the two batches due to the diversity in their origins.

With the data obtained in this manner, we have the opportunity to address our research question: Has fine-tuning the multilingual BERT model with Ottoman Turkish data facilitated the manual annotation and correction phase? To answer this question, we did the experiment using the two batches of annotated Ottoman Turkish sentences.

%In other words, we can now address the question of whether it is sufficient to fine-tune solely with modern Turkish, or if incorporating a small amount of Ottoman Turkish data for further fine-tuning results in annotations with fewer errors, thus making the correction easier for the annotators.

\subsection{Evaluation}

\begin{figure*}[t]
\centering
\includegraphics[width=\textwidth]{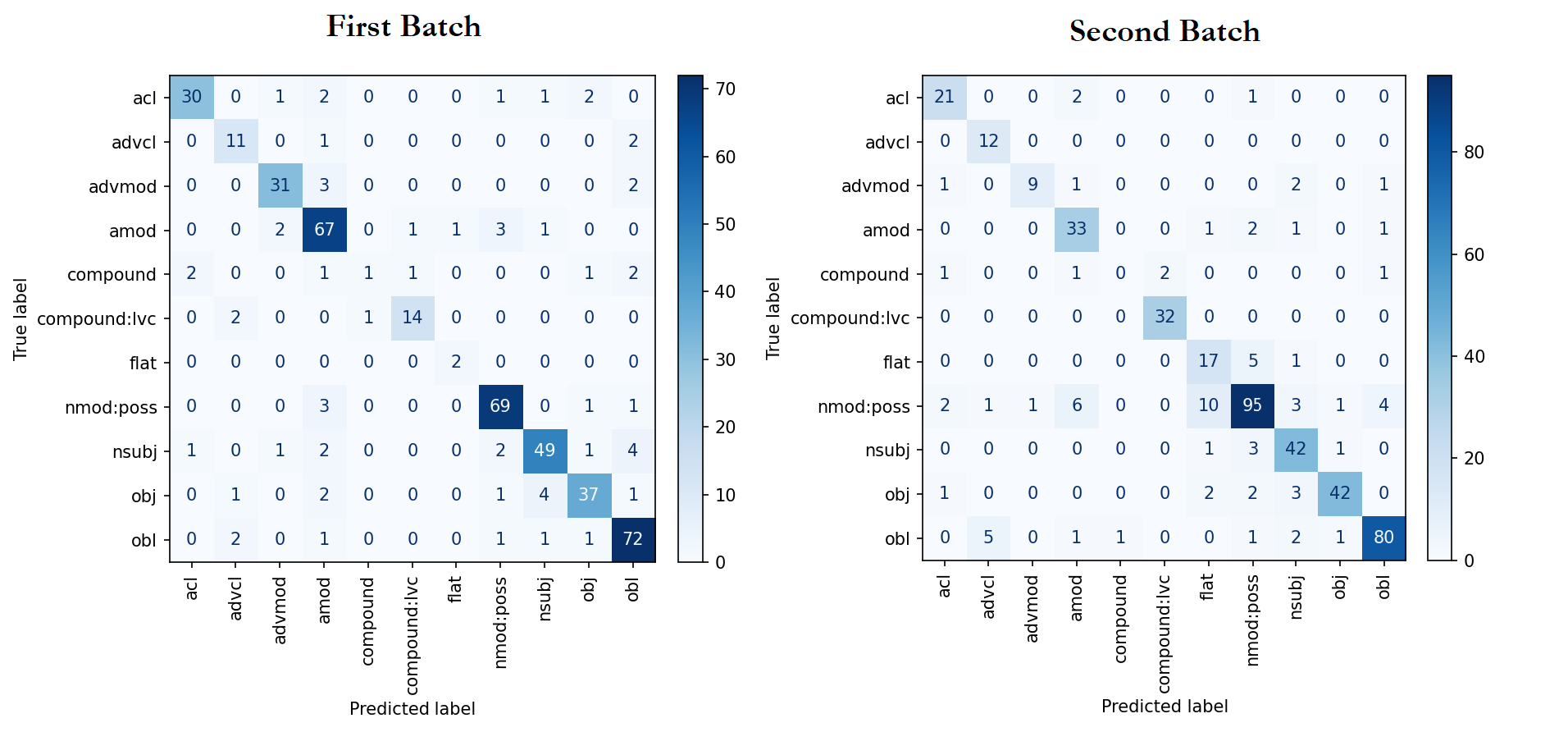}
\caption{Confusion matrices of gold and predicted dependency types on the first batch and the second batch. The x-axis in each plot shows the dependency types in the pseudo-annotations of the corresponding batch. The y-axis shows the dependency types in the gold annotations. }
\label{fig_cm}
\end{figure*}

In the experiment performed, we hypothesize that the difference in the annotations between `Second-Batch-Pseudo-Annotated' and `Second-Batch-Gold' should be less than the difference between `First-Batch-Pseudo-Annotated' and `First-Batch-Gold.' In other words, there should be fewer erroneous dependency relations to correct for human annotators in the second batch compared to the first batch.

\begin{table}

\centering
\begin{tabular}{lcc}
\hline
 & \textbf{First} & \textbf{Second}\\
  & \textbf{Batch} & \textbf{Batch}\\
\hline
\bf Size (\# of sentences) & 50 & 50  \\
\bf Ave. word count & 17.84 & \bf 20.45\\
%\bf \# of unique word forms & 569 & 524 \\
\bf UAS & 81.64 & \bf 82.58  \\
\bf LAS	& \bf 77.65	& 76.19 \\
\hline
\end{tabular}
\caption{\label{eva-res}
Comparison of the two batches in terms of unlabelled and labelled attachment scores.
}
\end{table}

To test this hypothesis, we measure the unlabeled and labeled attachment scores (UAS and LAS, respectively) between `First-Batch-Pseudo-Annotated' and `First-Batch-Gold' and between `Second-Batch-Pseudo-Annotated' and `Second-Batch-Gold'. Table \ref{eva-res} depicts the UAS and LAS F-1 scores of these measurements.

In Table \ref{eva-res}, we observe that further fine-tuning the BERT-based model with Ottoman gold-annotated data increases the success of the model in accurately predicting dependency arcs. However, there is a performance drop in correctly identifying the types of dependency relations.

\subsection{Quantitative Analysis}

To understand the performance drop in dependency type prediction of the updated parsing model, which was fine-tuned with Ottoman Turkish data, we compute confusion matrices for the first and second batches separately in Figure \ref{fig_cm}.

We observe that in the second batch there are many more {\textit{nmod:poss}} relations compared to the first batch. When we investigate the data, we see that a significant number of {\textit{nmod:poss}} relations in the second batch consist of Persian grammatical structures, which pose several challenges to annotation (see Section \ref{sec:annot-challenge}). Such structures are almost not present in the first batch. To be precise, in the second batch, approximately 20\% of the noun phrases are constructed using Persian grammar, whereas this figure is around 3\% in the first batch. The number of light-verb compounds is also higher in the second batch.

In addition to this, when we take into account the average word count in a sentence given in Table \ref{eva-res}, which is also higher in the second batch, we can conclude that annotating the second batch is a more challenging task for both the parsing model and human annotators. This might explain the slight performance drop in the LAS of the model when predicting the dependency types of the second batch.

Another factor that could affect the performance of the parsing model is the size of the fine-tuning data. The first batch is pseudo-annotated with the parser that is fine-tuned using BOUN, which has 7,803 annotated modern Turkish sentences in its training set. The second batch is pseudo-annotated after this parser is further fine-tuned with only 50 annotated Ottoman Turkish sentences. Such a small amount of data may not be sufficient for effectively fine-tuning the model. We anticipate that as we annotate new batches using this iterative annotation method, the performance of the parser will improve progressively.

Based on the results of the first and second batch runs, we can say that our iterative approach helps understand the difficulties of annotating Ottoman texts layer by layer. The first batch of fine-tuning process reveals mostly lexical issues. The results of the second batch of fine-tuning point to the complex grammatical issues that can be quite prevalent in certain types of texts.

\subsection{Conclusion}

Generation of a treebank for Ottoman Turkish is a difficult process that requires expertise in more than one language. The defunct grammatical forms and obsolete vocabulary make manual annotation a challenging task. We present an iterative approach utilizing a pretrained large language model, multilingual BERT, for annotating the first Ottoman Turkish dependency treebank. Our empirical findings suggest that performing the manual data annotation iteratively in a human-in-the-loop fashion improves and eases the process of dependency annotation. We anticipate that, when completed, the resulting treebank will enhance the NLP of Ottoman Turkish texts and enable a more profound exploration of Ottoman Turkish linguistic and cultural nuances.

% Entries for the entire Anthology, followed by custom entries
\bibliography{acl_latex.bib}

\end{document}